\documentclass[letterpaper]{article} 
\usepackage[submission]{aaai25}  
\usepackage{times}  
\usepackage{helvet}  
\usepackage{courier}  
\usepackage[hyphens]{url}  
\usepackage{graphicx} 
\usepackage{natbib}  
\usepackage{caption} 
\usepackage{algorithm}
\usepackage{algorithmic}

\usepackage{newfloat}
\usepackage{listings}
\DeclareCaptionStyle{ruled}{labelfont=normalfont,labelsep=colon,strut=off} 
\floatstyle{ruled}
\newfloat{listing}{tb}{lst}{}
\floatname{listing}{Listing}
\title{Planning Transformer: Long-Horizon Offline Reinforcement Learning with
Planning Tokens}
\author {
    Joseph Clinton\textsuperscript{\rm 1}, 
    Robert Lieck\textsuperscript{\rm 2}
}

\affiliations {
    \textsuperscript{\rm 1}Independent Researcher (formerly at Durham University), 
    \textsuperscript{\rm 2}Durham University\\
    joeclinton1@btinternet.com, robert.lieck@durham.ac.uk
}

\usepackage{xcolor}

\usepackage{caption}
\usepackage{tabularray}
\usepackage{booktabs}
\usepackage[shortcuts]{glossaries-extra}
\glsdisablehyper 
\setabbreviationstyle[acronym]{long-short}  
\GlsXtrEnableEntryCounting{acronym}{3}  
\MFUhyphentrue 
\newacronym[longplural=returns-to-go]{rtg}{RTG}{return-to-go}
\newacronym{dt}{DT}{Decision Transformer}

\begin{document}

\maketitle

\begin{abstract}
Supervised learning approaches to offline reinforcement learning, particularly those utilizing the Decision Transformer, have shown effectiveness in continuous environments and for sparse rewards. However, they often struggle with long-horizon tasks due to the high compounding error of auto-regressive models. To overcome this limitation, we go beyond next-token prediction and introduce Planning Tokens, which contain high-level, long time-scale information about the agent's future. Predicting dual time-scale tokens at regular intervals enables our model to use these long-horizon Planning Tokens as a form of implicit planning to guide its low-level policy and reduce compounding error. This architectural modification significantly enhances performance on long-horizon tasks, establishing a new state-of-the-art in complex D4RL environments. Additionally, we demonstrate that Planning Tokens improve the interpretability of the model's policy through the interpretable plan visualisations and attention map.
\end{abstract}

%

\section{Introduction}

Offline reinforcement learning (Offline RL) has emerged as a powerful paradigm, enabling agents to learn effective policies from a fixed dataset without requiring interaction with the environment, making them particularly beneficial in scenarios where data collection is costly or unsafe \cite{levineofflinerl_tutorial}.

A recent paradigm shift in Offline RL has come in the form of Reinforcement learning via supervised learning (RvS) which approaches Offline RL as a sequence modelling problem, with one notable implementation of this framework being the Decision Transformer \cite{DT}.  Unlike with traditional temporal difference based Offline RL methods, \ac{dt}s perform credit assignment directly, making them highly sample efficient. They are resilient to distractor signals, excelling in sparse reward environments, and can successfully model multimodal distributions, enabling better generalization and transfer across different tasks \cite{DT, TT, sampleefficientworldmodels}.

Despite these advantages, RvS methods, including  \ac{dt}s, still face two significant challenges:
\begin{enumerate}
\item \textbf{Compounding Error:} Auto-regressive token prediction suffers from compounding error in long, complex, or multi-task environments. RL environments, unlike language, are very sensitive to small errors as without opportunities for "checkpoints" (i.e., points where the agent can re-calibrate), even small errors will accumulate along the trajectory. \cite{TT, compounding_error}.
\item \textbf{Credit Assignment:} Whilst \ac{dt}s can handle credit assignment directly to an extent, they still require long contexts to assign credit effectively over long horizons. Without these contexts, which can lead to increased model complexity and slower training, their ability to reinforce an optimal long-term policy is limited \cite{WT}.
\end{enumerate}

Hierarchical reinforcement learning (HRL) offers a solution by decomposing long tasks into a series of shorter, more manageable sub-tasks. HRL employs a high-level controller to select sub-tasks or subgoals for a low-level worker policy, facilitating more efficient learning and reducing the impact of compounding error. However, HRL models introduce greater complexity, making them harder to train and extend. They are also task-specific and can even worsen the credit assignment problem due to their distinct separation between the low and high level tasks. \cite{HRL-survey}.

In this paper, we introduce a novel agent architecture that combines the strengths of RvS and HRL. We extend the  \ac{dt} framework by incorporating high-level Plans that the agent can learn to generate and use, implicitly guiding its long-horizon decision-making. This hybrid approach leverages the hierarchical decomposition of tasks to manage compounding errors, whilst maintaining the simplicity and efficiency of RvS. By incorporating the Plans seamlessly with the RvS trajectories and using a unified model, we additionally overcome HRL's credit assignment limitation as there is no explicit distinction between the high and low level policies.

Our main contributions are as follows:
\begin{enumerate}
    \item \textbf{Dual-Timescale Token Prediction (Planning Tokens)}: We introduce the Planning Transformer model, that uses the novel concept of extending RvS methods with High-Level Planning Tokens. 
    \item \textbf{State-of-the-Art Offline-RL Performance}: We demonstrate our approach is competitive with, or exceeds state-of-the-art offline-RL methods in both long and short horizon tasks,
    \item \textbf{Advancements in Long-Horizon Interpretability}:  We show our Plans increase the interpretability of the long-horizon decision-making processes within an RL agent.
\end{enumerate}

\section{Related works}

Our paper integrates Offline RL, Hierarchical RL, and Model-based planning. We'll briefly explore each domain and relevant literature.

\subsection{Offline RL via Supervised Learning (RvS)}

Offline RL learns action policies from fixed datasets, valuable when online data collection is costly or unsafe. Challenges include sparse rewards and out-of-distribution states. RvS, introduced with Trajectory Transformer \cite{TT} and Decision Transformer \cite{DT}, addresses these by modeling Offline RL as a sequence problem. This approach naturally adheres to in-distribution actions and performs well in sparse reward environments.

RvS excels in sparse reward settings but underperforms Temporal difference methods like CQL \cite{CQL} and IQL \cite{IQL} in dense reward environments. It struggles with long-horizon tasks due to compounding error \cite{compounding_error} and difficulty in assigning value to states \cite{IQL, WT}. While possible with a two-layer FNN \cite{rvs}, RvS commonly uses transformers, which are expensive to train \cite{RT-X, transformers_lighter_survey}.

Our model's Planning Tokens enables it to directly address the compounding error and credit assignment problem, enhance overall performance even in dense reward environments, and achieve strong results with few parameters making training faster and cheaper.

\subsection{Hierarchical RL}

HRL methods tackle long-horizon decision-making by learning high-level policies for selecting subgoals or subtasks, while low-level policies execute these actions \cite{Option-Critic, FuN, HIRO, LEXA, DADS, STRAW}. Key models include the option-critic and feudal network architectures.

Recent advancements include automated skill discovery \cite{LEXA, DADS, STRAW} and using latent space world-models for goal sampling \cite{implicitrlatscale, TAP}. Large language models (LLMs) have been explored as high-level controllers for proposing subgoals \cite{Voyager, Language-Planner} or subtasks \cite{SayCan,RT-1,PaLME-E}.

Despite progress, HRL methods face challenges in skill discovery and maintaining coherence between high and low-level policies. Our method addresses these issues, integrating strengths from both feudal and option-critic frameworks.

\subsection{Model-Based Planning}

Model-based Planning (MDP) uses forward dynamics models to predict future states, which in turn are used to improve state-value estimates for Offline-RL. This approach excels in high-dimensional environments with sparse rewards, where direct learning is challenging.

MBP has been applied to discrete long-horizon tasks using Monte Carlo Tree Search (MCTS) methods \cite{dynaq, alphago, alphazero, MuZero} where the model performs rollouts in perfect information environments with small action spaces. MBP has also been extended to continuous search spaces by combining cross-entropy methods with probabilistic forward dynamics models often within latent observation spaces \cite{PETS, P2E, DreamerV1, worldmodels}. Recent research has explored folding the forward-dynamics planning model into the action policy through iterative denoising diffusion policies \cite{Diffuser, DiffusionPolicy, DiffusionQL}, as-well as learning latent-temporal spaces for efficient compact planning \cite{TAP, SeCTAR, vq-plan, PlaNet}

Our method is similar to MDP in that our Plan Generator is a forwards dynamics model that guides our action policy, however notably we do not perform a search within this space it is only used for guidance to the action policy. Our Plans are also in a latent planning space, however our method does not require an autoencoder to learn this space. Like diffusion policy methods our planning and action policy are unified however our method uses a transformer backbone allowing it to remain auto regressive reducing latency and allowing it to adapt to environmental changes.

\subsection{Hierarchical Decision Transformers}

Three principal models address long-horizon challenges in DT through Hierarchical RL: HDT \cite{HDT}, WT \cite{WT}, and ADT \cite{ADT}. Each uses a high-level policy to propose goal prompts for a decision-transformer acting as a low-level policy.

HDT uses a hierarchical sub-goal policy with independent high and low-level DTs. WT employs a simple FNN for goal prediction, combining state and goal tokens. ADT uses HIQL \cite{HIQL} for goal prediction, achieving SOTA results on the D4RL benchmark.

Our model introduces Multi-token Plans, flexible conditioning targets, fixed-interval re-planning, and a unified model. This approach surpasses prior works in accuracy, flexibility, efficiency, and simplicity, while offering interpretability.

\section{Method}

The original \ac{dt} \cite{DT} predicts the next action using a GPT-style Transformer \cite{transformer, gpt2} that takes as input the previous \(\tau\) \acp{rtg}, states, and actions. We denote \acp{rtg} with \(r\), states with \(s\) and actions with \(a\):
\begin{equation}
    (r_0, s_0, a_0, r_1, s_1, a_1, \dots, r_{|\tau|}, s_{|\tau|}, a_{|\tau|})
\end{equation}

We propose to extend \ac{dt} with a Planning head which predicts \(n\) Planning Tokens that contain information about the agents future states, actions and \acp{rtg}, and then pre-pending these to its input sequence, so that its next action prediction is conditioned on its very own prediction of its future. We denote sequences of these Planning Tokens as Plans.

This simple modification effectively reduces the action-horizon from long to short. Additionally it allows a model to learn how its short term actions, affect its long term success without needing to provide the entire trajectory during training and inference.

For the remainder of this section we will explore how Plans are sampled from the trajectory, how we incorporate these Plans and how we train and evaluate our model.

\subsection{Plan Representation}

\subsubsection{Plan Sampling}

A Plan is a temporally high level representation of a trajectory, as such prior works have generated Plans by temporally compressing trajectories within the dataset using Variational Autoencoders  \cite{TAP, PlaNet, SeCTAR}. We opt for a simpler method of just sparsely selecting timesteps from throughout the trajectory and concatenate them together.

This would seem to lose necessary information about the trajectory and one would expect it to be ineffective but we find quite the opposite. We believe that the reason our method remains effective despite its simplicity is that our unified training method means that naturally the model learns to optimize for Plans that contain information that is beneficial to the action prediction policy. This means that despite the Plans being simple they remain effective for guidance to the action policy.

We explored various methods for how to sample these timesteps, which we explore in detail in section \ref{ablation}.

\subsubsection{Planning Token representation} \label{representation}

A Planning Token is a subgoal containing information about the agent's long-horizon future trajectory. Each Planning Token:
\begin{itemize}
    \item Maps to one future time step
    \item Contains the full observation feature space or subset of it
    \item Contains \acp{rtg} targets for reward conditioned environments
    \item May contain the corresponding Action for that observation.
\end{itemize}

We explore various design choices for this representation in Section \ref{ablation}

\subsection{Input Sequence Construction}

Once we have constructed the Plans we pair them with our trajectories during the dataset batch loader process.

Our Sequence Construction is described as follows:

\begin{enumerate}
    \item Plans are inserted after the first state and first return-to-go and before the first action
    \item We subtract the first state of the plan during training or current state observation during evaluation to make the Plans relative rather than absolute.
\end{enumerate}

Our adjusted input sequence for PT is now:

\begin{equation}
\begin{split}
      &(r_0, s_0,\\
    &p_0-s_0, p_1-s_0, \ldots, p_n-s_0, \\
    &a_0 r_1, s_1, a_1, \ldots, r_{|\tau|}, s_{|\tau|}, a_{|\tau|})
\end{split}
\end{equation}

\subsubsection{Goal Conditioning} \label{goal-conditioned}

We use target observations as goal-conditioning, and insert these targets as the first token in the input sequence in order to condition both the plan and action policy on it. For simplicity, during training our goals are selected as the last observation in the Plan. 

We use a novel goal representation where we project the first state into goal space and concatenate it to the goal. Our input-space adjusted for goal-conditioning is now:

\begin{equation}
\begin{split}
      &(\text{proj}_g(s_0), g, r_0, s_0,\\
    &p_0-s_0, p_1-s_0, \ldots, p_n-s_0, \\
    &a_0 r_1, s_1, a_1, \ldots, r_{|\tau|}, s_{|\tau|}, a_{|\tau|})
\end{split}
\end{equation}
where $\text{proj}_g(s_0)$ denotes the projection of the first state $s_0$ into the goal space $g$.

\begin{figure}
\centering
\includegraphics[width=1\linewidth]{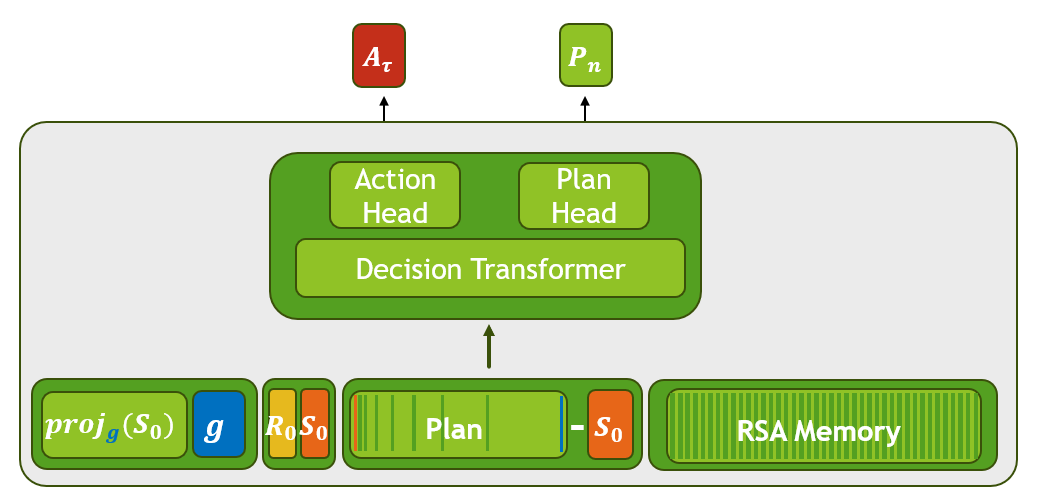}
\caption{Planning Transformer Architecture Diagram showing the input sequence construction and the two output heads. RSA memory denotes the agent's memory of the more recent \acp{rtg}, states, actions.}
\label{fig:architecture}
\end{figure}

We explored alternative goal representations in section \ref{ablation}. Figure \ref{fig:architecture} shows the input sequence and a high level simplification of the model architecture.

\begin{table*}
\caption{Normalised scores per task on Mujoco, Antmaze and FrankaKitchen environments, where \textbf{bold} highlighting indicates SOTA performance within either the non-RvS or RvS category. The vertical line separates non-RvS and RvS based methods respectively.}
\centering
{\footnotesize
\begin{tblr}{
width = \linewidth,
colsep = 4pt,
colspec = {X[l] c c c c c c c c},
cell{1}{2,3,4,6,7,8,9} = {c},
cell{2-5}{2,3,4,6} = {c},
cell{5}{7,8,9} = {c},
cell{13-15}{2,3,4,6} = {c},
vline{5} = {-}{},
hline{1-2,8,12,14} = {-}{},
}
\textbf{Environment}   & \textbf{CQL}             & \textbf{IQL}             & \textbf{HIQL}           & \textbf{RvS-R/G} & \textbf{DT}      & \textbf{WT}             & \textbf{G/V-ADT}        & \textbf{PT (OURS)}      \\
halfcheetah-replay     & \textbf{45.5~$\pm$ 0.5}  & 44.2~$\pm$ 1.2           &        -                & 38.0$\pm$ 0.7    & 35.4$\pm$~1.6    & 39.7$\pm$ 0.3           & \textbf{42.8$\pm$ 0.2}  & 41.3$\pm$~1.0            \\
hopper-replay          & \textbf{95.0$\pm$~6.4}   & 94.7~$\pm$ 8.6           &           -              & 73.5$\pm$ 12.8   & 43.3$\pm$ 23.9   & 88.9$\pm$ 2.4           & 83.5$\pm$ 9.5           & \textbf{89.7 $\pm$~4.5}   \\
walker2d-replay        & \textbf{77.2~$\pm$ 5.5}  & 73.8~$\pm$ 7.1           &           -              & 60.5$\pm$ 6.7    & 58.9$\pm$ 7.1    & 67.9$\pm$ 3.4           & \textbf{86.3$\pm$ 1.4}  & 55.5$\pm$~6.2            \\
halfcheetah-expert     & \textbf{91.6~$\pm$ 2.8}  & 86.7~$\pm$ 5.3           &            -             & 92.2$\pm$ 1.2    & 84.9~$\pm$ 1.6   & \textbf{93.2}$\pm$~0.5           & 91.7$\pm$ 1.5           & 92.6$\pm$ 0.8   \\
hopper--expert         & \textbf{105.4~$\pm$ 6.8} & 91.5~$\pm$ 14.3          &             -            & 101.7$\pm$ 16.5  & 100.6$\pm$ 8.3   & \textbf{110.9$\pm$ 0.6} & 101.6$\pm$ 5.4          & 109.9$\pm$ 1.1           \\
walker2d-expert        & 108.8~$\pm$ 0.7          & \textbf{109.6~$\pm$ 1.0} &          -               & 106.0$\pm$ 0.9   & 89.6$\pm$ 38.4   & 109.6$\pm$ 1.0          & \textbf{112.1$\pm$ 0.4} & 108.8$\pm$~0.9       \\
antmaze-umaze-diverse  & 84.0                     & 62.2~$\pm$ 13.8          & \textbf{87.6~$\pm$~4.8} & 60.9$\pm$ 2.5    & 42.2$\pm$ 5.4    & 71.5$\pm$ 7.6           & 83.8$\pm$ 2.3           & \textbf{86.6$\pm$ 4.3}   \\
antmaze-medium-diverse & 53.7                     & 70.0$\pm$~10.9           & \textbf{87.0~$\pm$~8.4} & 67.3$\pm$ 8.0    & 0.0$\pm$ 0.0     & 66.7$\pm$ 3.9           & 83.4$\pm$ 1.9           & \textbf{85.4$\pm$ 2.1}   \\
antmaze-large-diverse  & 14.9                     & 47.5~$\pm$ 9.5           & \textbf{81.2~$\pm$~6.6} & 36.9$\pm$ 4.8    & 0.0$\pm$ 0.0     & 72.0$\pm$ 3.4           & 65.4$\pm$ 4.9           & \textbf{82.3~$\pm$~5.8}  \\
antmaze-ultra-diverse  & 9.6~$\pm$ 14.6                    & 17.8~$\pm$ 9.5           & \textbf{52.9~$\pm$~17.4} & 26.4~$\pm$7.7   & 0.0$\pm$ 0.0     & -           & -           & \textbf{34.9 ~$\pm$~7.1}  \\
kitchen-partial        & 49.8                     & 46.3                     & \textbf{65.0~$\pm$~9.2} & 51.4$\pm$ 2.6    & 31.4$\pm$ 19.5   & 63.8$\pm$ 3.5           & 64.2$\pm$ 5.1           & \textbf{66.7~$\pm$ 5.0}  \\
kitchen-mixed          & 51.0                     & 51.0                     & \textbf{67.7$\pm$~6.8}  & 60.3$\pm$ 9.4    & 25.8$\pm$ 5.0    & 70.9$\pm$ 2.1  & 69.2$\pm$ 3.3           & \textbf{71.3$\pm$~2.5}            \\  
\end{tblr}
}
\label{Tab:results}
\end{table*}

\subsection{Unified Training and Inference Pipeline}
\subsubsection{Unified training}

The architecture of our model remains largely consistent with the original  \ac{dt}. The primary modification is the addition of a planning head, which takes as input \(K\) consecutive tokens from the input sequence starting from \(s_0\) and outputs \(K\) corresponding planning tokens.

To train this modified model, we employ a combined loss function:
\[
\mathcal{L} = \alpha \cdot \mathcal{L}_{action} + \beta \cdot \mathcal{L}_{plan}
\]
Here, \(\mathcal{L}_{action}\) represents the L2 norm action loss, and \(\mathcal{L}_{plan}\) is a newly introduced L2 norm plan deviation loss. 

The value of \(\alpha\) and \(\beta\) depends on the feature size of the Plans and the actions. In most cases we use \(\alpha = \beta = 0.5\) to balance the two, but it may be necessary to bias the optimizer towards one of the other if one policy is significantly harder to learn than the other.

\subsubsection{Inference Pipeline}

Inference using the model follows the standard procedure of \ac{dt}, however before proceeding with the autoregressive generation loop and environment query we first generate a plan. This is achieved by providing PT with \(r_0\), \(s_0\), and \(g_0\) for goal-conditioned inputs, and then querying the PT model \(n\) times. In each iteration \(i\), we select the token at index \(i\) from the plan head's output and insert it into the initially all-zero plan sequence at index \(i\).

Once the plan is generated, it is fed into the PT model along with the states, actions, and \acp{rtg}. Additionally, as described in Section \ref{goal-conditioned}, we also input the goal for goal-conditioned environments.

Every \(\rho\) timesteps, we regenerate the plan using the state from the last timestep as the new initial state, ensuring proper replanning. We found replanning more frequently improves accuracy at the cost of computational efficiency so we use \(10<\rho<50\).

\begin{figure}
\centering
\includegraphics[width=1\linewidth]{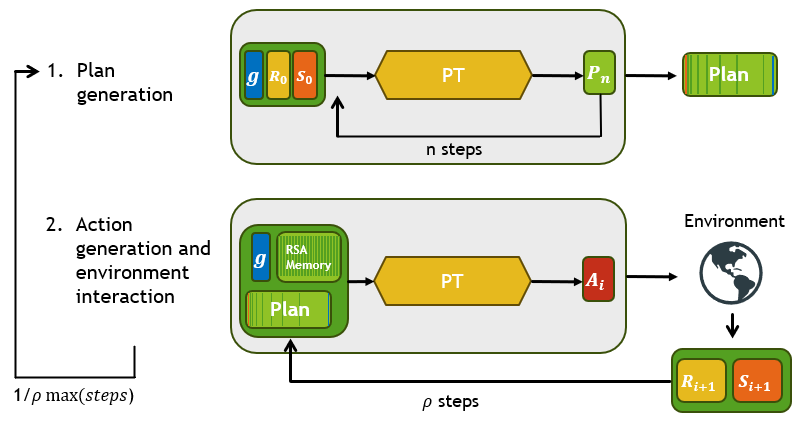}
\caption{Model Inference pipeline. In Stage 1 we use PDT's planning head to generate $\rho$ planning tokens, and in Stage 2 we switch to the action head to generate the actions for the agent conditioned on the Plan  }
\label{fig:inference}
\end{figure}

See Figure \ref{fig:inference} for an overview of the model inference pipeline.

\section{Evaluation}

In our experiments, we evaluated our methods across various environments to comprehensively assess their performance under different conditions.

For \textbf{goal-conditioned environments}, we tested our methods on both short and long-horizon tasks to evaluate the model's ability to achieve specified objectives in diverse scenarios. The \textbf{AntMaze} environment, where an 8-DOF quadruped ("Ant") navigates mazes of varying sizes (umaze, medium, large, ultra), was particularly useful for assessing the model's "trajectory stitching" capabilities—its ability to connect suboptimal trajectories to reach the desired goals. Additionally, the \textbf{FrankaKitchen} environment featured a 9-DOF Franka robot performing a sequence of goal-conditioned tasks in a kitchen setting.

For \textbf{reward-conditioned tasks}, we used the \textbf{Gym-Mujoco} suite, focusing on locomotion and control tasks with varying trajectory quality. We tested in \textbf{medium-replay} (medium + low-quality trajectories) and \textbf{medium-expert} (medium + high-quality trajectories) settings.  The environments are \textbf{HalfCheetah} (a multi-jointed robot running forward), \textbf{Hopper} (a one-legged robot hopping without falling), and \textbf{Walker2d} (a bipedal robot balancing and walking). 

We maintain mostly consistent hyperparameters across all environments and tasks, however we vary whether the model uses timestep embedding, the sequence length and embedding dropout probability for certain environments. We detail our hyperparameters in more detail in the supplementary material.

\subsection{Ablation Study} \label{ablation}

\begin{table*}[h]
\caption{Combined ablation test results for different parameters across D4RL environments.}
\begin{tblr}{
  width = \linewidth,
  colspec = {X[c] X[c] X[c] X[c]}, 
  hline{1, Z} = {1pt},
  hline{2-0} = {0.5pt},
}

\textbf{Value} & \textbf{Kitchen-Mixed} & \textbf{Antmaze-Large} & \textbf{Hopper-Replay} \\

\hline
\SetCell[c=4]{c}\textbf{Plan Sampling Method} \\
\hline
Fixed-time & 60.1 & 47.4 & 84.8 \\
Fixed-distance & \textbf{71.3} & 81.3 & \textbf{89.7} \\
Log-time & 37.9 & 60.6 & 87.3 \\
Log-distance & 32.9 & \textbf{82.3} & 89.4 \\

\hline
\SetCell[c=4]{c}\textbf{Plan Use Relative States} \\
\hline
Relative States & \textbf{71.3} & \textbf{82.3} & \textbf{89.7} \\
Absolute States & 42.7 & 51.6 & 57.3 \\

\hline
\SetCell[c=4]{c}\textbf{Goal Representation} \\
\hline
Absolute Goal & 61.3 & 69.9 & 74.8 \\
Relative Goal & 64.9 & 75.2 & 74.5 \\
Project State to Goal and Absolute Goal & \textbf{71.3} & \textbf{82.3} & \textbf{89.7} \\
Project State to Goal and Relative Goal & 44.8 & 81.1 & 75.4 \\

\hline
\SetCell[c=4]{c}\textbf{Use actions in Plan} \\
\hline
True & \textbf{71.3} & 71.1 & 84.0 \\
False & 52.3 & \textbf{82.3} & \textbf{89.7}\\
\end{tblr}
\label{table:combined_ablation_tests}
\end{table*}

\subsubsection{Plan Sampling Method}
We tested four different methods for how to sample Plans: fixed-timestep width, fixed-distance width, and logarithmic-distance sampling. Fixed step sampling refers to sampling at equal gaps of either timesteps or distance, whilst logarithmic sampling, samples more timesteps from early on in the trajectory than later.

We found that distance-based sampling often produced better performance which we hypothesize is because it provides Plans which are more information-rich. We found that for ant-maze log sampling was more effective, for hopper replay it was a little less effective, but surprisingly it was much less effective for kitchen. Fixed-distance seems to be most overall effective.

\subsubsection{Plan Use Relative States}
Using relative states consistently improved performance across all environments compared to absolute states. We hypothesize that planning in relative space, helps the model generalize its planning policy.

\subsubsection{Goal Representation}

We tested goals being absolute observation space, goals being in relative observation space. Aswell as two more experimental ideas, of projecting the first state to the goal space and concatenating it with the goal in either absolute or relative observation space.

Generally, the experimental approach of projecting the first state to goal space and concatenating it to the goal, out performed more naive methods, with the absolute goal version doing best. Our observations indicate that relative goals, helped the model generalize but decreased the accuracy of the Plans, and that absolute goals did the opposite. So we hypothesize that this goal representation lets the model generalize whilst keeping the goals accurate.

\subsubsection{Use actions in plan}

We find that whether or not to use actions in the policy is not clear, as it benefits some environments like Kitchen but in others it hurts performance. We hypothesize that actions can make the planning policy harder to learn, and may provide little additional information in some environments, but in environments with complex action spaces like Kitchen, inlcuding actions can give the Planning Tokens sub-task qualities aswell as sub-goal qualities which might be beneficial. 

\subsection{Comparison with Prior Methods}

Our evaluation benchmarks the performance of PT against various state-of-the-art (SOTA) Offline RL methodologies. We include the following:

\begin{itemize}
    \item \textbf{CQL} \cite{CQL} and \textbf{IQL} \cite{IQL}: Examples of SOTA Offline RL methods.
    \item \textbf{HIQL} \cite{HIQL}: An example of a SOTA goal-conditioned Offline RL method.
    \item \textbf{RvS-R/G} \cite{rvs} and \textbf{DT} \cite{DT}: Baselines for RvS methods, with DT being the model we extended.
    \item \textbf{WT} \cite{WT} and \textbf{G/V-ADT} \cite{ADT}: Examples of SOTA goal-conditioned  \ac{dt} variants.
\end{itemize}

For all methods, we reference reported results from previous works \cite{ADT, WT, goalconditionedpredictivecoding, HIQL}. Except for PT, scores are reported using the evaluation methodology from \cite{rvs}, where a score is the average of the scores from 5 random training seeds with 100 rollouts each on the last checkpoint. In the case of PT, we used 3 random training seeds instead of 5 due to the limited computational resources of our lab. However, the low standard deviation of our results across 3 seeds suggests that significant deviations are unlikely even if we had used the full 5 seeds.
\\~\\
Table \ref{Tab:results} presents our results on the D4RL evaluation benchmarks compared to prior methods. The primary findings are as follows:

\begin{enumerate}
    \item Our main result is that on the goal-conditioned long-horizon benchmarks, Antmaze and FrankaKitchen, our model \textbf{outperforms  all RvS methods}, and outperforms Offline-RL methods in general on 3/6 environments.
    \item Even in reward-conditioned environments where, due to the lack of clear sub-goals, one might expect Planning Tokens to be ineffective, they still present a notable improvement upon no-plans. Our model is \textbf{competitive with SOTA} in  5/6 environments and surpasses SOTA on 1 environment
\end{enumerate}

Our SOTA results on the long-horizon goal-conditioned Antmaze and FrankaKitchen environment prove empirically that our novel concept of planning-tokens are effective in long-horizon environments.

Our competitive with SOTA results on the reward conditioned MuJoCo environments show that our Plans demonstrate applicability to both reward conditioned and goal-conditioned environments. These results are more impressive when considering that CQL, IQL, HIQL, WT and V-ADT use architectures specific to dense rewards environments, however our model does not explicitly target reward conditioned environments.

Additionally, our method is considerably simpler and more flexible, and more interpretable than most Offline-RL methods. Therefore, these advantages may warrant consideration even in scenarios where the model does not achieve SOTA performance.

\subsection{The Utility and Interpretability of Planning with PT}

\subsubsection{Utility of the Plans}

\begin{figure}
    \centering
    \includegraphics[width=1\linewidth]{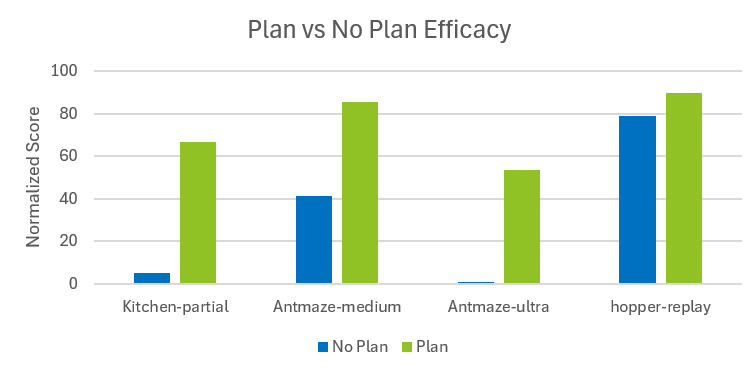}
    \caption{Normalized score of model on a selection of environments with Plans and without Plans, demonstrating the efficacy of using Plans.}
    \label{fig:with-without-plans}
\end{figure}

Due to our addition of goal-conditioning and improved hyper-parameter tuning, it may be unclear to what degree our results are due to the Planning Tokens addition on their own.
To demonstrate the efficacy of our planning approach, in Figure \ref{fig:with-without-plans} we compare the normalized score of the model with Plans and with Plans disabled. Without Plans, the goal-conditioned transformer outperforms DT on many environments, for example on AntMaze-Medium-Diverse it achieves 41 which is much higher than DT's score of 0.0. However, for all environments, using Plans increases performance, over the No-Plan variation. The degree of improvement in the models performance appears dependent on the complexity of the environment as it significantly enhances performance in long-horizon in  long-horizon environments like Kitchen-Partial but only slightly enhances performance in short-horizon environments like hopper-replay. Overall, this clearly demonstrates that the superior performance of PT is due to its Plans, not its goal-conditioning or other minor enhancements.

\subsubsection{Interpretability of the Plans}

\begin{figure}
    \centering
    \includegraphics[width=1\linewidth]{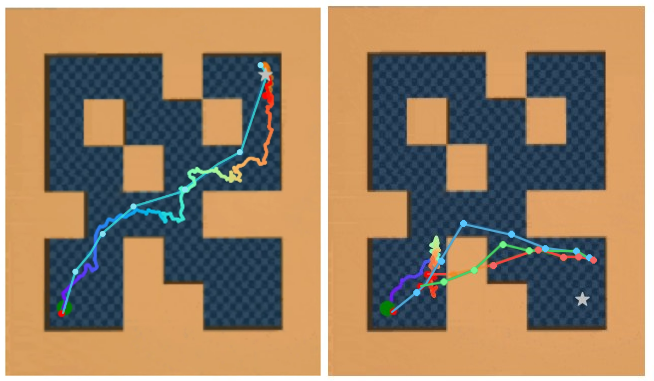}
    \caption{Visualisation of Plans on antmaze-large environment. The rainbow coloured line is the actual path taken by the ant, the star is the goal, and the solid coloured lines with key-points (beginning at the red dots) are the Plans. On the left is a success case where the agent reached the goal and on the right is a failure case where the model got stuck. In both cases the Plans provide interpretability.}
    \label{fig:plan-visualisation}
\end{figure}

By projecting our Plans into a 2D space we can visualise our agent's high level thinking as it Plans where to go next. This could potentially introduce a significant enhancement in the interpretability of reinforcement learning models. The left image in Figure \ref{fig:plan-visualisation} shows an example of one of these visualisations on AntMaze large, based on a Plan made by the agent at the beginning of the trajectory. From the figure we can see the model's plan matches closely with its eventual trajectory. The right image in Figure \ref{fig:plan-visualisation}, shows an example of a failure case where the agent failed to navigate around a wall. Normally, the opaque nature of the model, would make it challenging to identify the exact reason for the failure, but by visualising the Plans, we can see that the model has incorrectly inferred that the optimal path to the goal traverses the wall. In the real world, having advance knowledge of an agents high level policy before it carries it out, can be very useful for preventing the agent carrying out an unsafe policy, as theoretically the Plans high level and interpretable nature makes them well suited as input for a secondary safe guard model.

\begin{figure}
    \centering
    \includegraphics[width=0.75\linewidth]{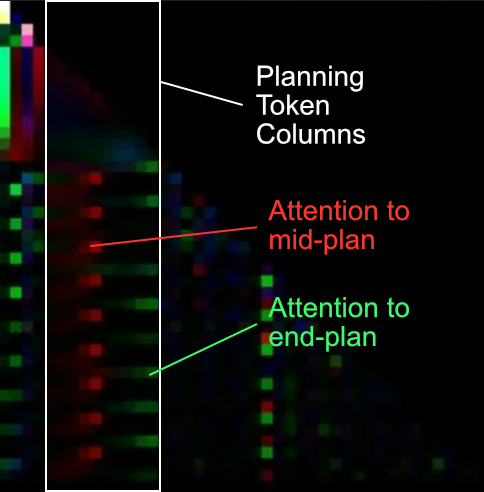}
    \caption{A snapshot of the attention map taken halfway through a run on the AntMaze medium-diverse environment.}
    \label{fig:attention}
\end{figure}

Not only do the Plan Visualisations, enhance interpretability but PT's attention maps, when focusing on the columns relevant to its Planning Tokens, also enhance interpretability. Figure \ref{fig:attention} shows an attention map where red, green, and blue represent layers 1,2,3 of the Transformers Attention heads respectively and columns 5-15 are the attention scores on the Planning-Tokens themselves. Planning Token attention is visualised in columns 5-15. This unified visualisation offers quick insights into the model's "thoughts." For example, the mid-Plan (columns 4-5) is highlighted in red, and the end-Plan (columns 9-10) in green. Since layer 1 (red) focuses on low-level and layer 2 (green) on medium-level information, we could potentially infer that the model has learnt to focus on the mid point of the plan as a lower level immediate goal but at a higher level is still guided towards reaching the final goal. This interpretability sets the PT model apart from HDT, ADT, and WT, and could be an interesting area of future research.

\section{Future Work}
The Planning Transformer (PT) introduces a robust framework that opens several promising avenues for future research, aiming to expand its applicability and efficacy.

An immediate area of interest is the expansion of PT into online learning environments. Building on the groundwork laid by the Online  \ac{dt} research \cite{ODT}, PT's algorithmic Plan generation is well-suited for online adaptation. As new data is collected through exploration, the Plan generation policy can be dynamically refined.

Addressing the limitations in non-Markov environments represents another critical direction for enhancing PT. Whilst sparse Plans have been surprisingly effective in the environments we benchmarked, it is possible there are environments where this simple method of generating Plans would not suffice, for example environments where decisions depend on intricate historical contexts. Incorporating temporal encoders, such as convolutional autoencoders or leveraging the Transformer architecture itself for dense temporal encoding, could significantly improve the representation and utility of Plans.

Finally, since PT is just a single transformer model, there is little preventing this framework from being applied to other domains that use transformers, such as NLP. Recently, it has been shown that multitoken prediction can actually improve an LLM's coding performance \cite{multitoken-llm-meta}, as such it would be particularly interesting to see if PT applied to LLMs would provide similar enhancements particularly in their long-horizon and complex reasoning abilities. 

\subsection{Conclusion}
We introduce the Planning Transformer (PT), a novel framework that integrates multi-token planning with transformer-based reinforcement learning via supervised learning methods. By prepending multi-token Plans to trajectory inputs, PT enhances long-horizon decision-making capabilities in Offline RL settings. Our approach achieves performance competitive or surpassing SOTA across a variety of challenging benchmarks in the D4RL Offline RL suite, while also being simpler and more flexible. The explicit incorporation of hierarchical planning enables better trajectory stitching and strategic reasoning over long time horizons. Furthermore, PT provides interpretability benefits by visualising the generated Plans and attention maps, promoting transparency in the model's decision-making process. Overall, our work represents a transformative integration of planning and transformers for Offline RL, opening new avenues for model-free hierarchical reinforcement learning. Future directions include extending PT to online adaptation, handling non-Markovian environments through temporal encoders, and incorporating the framework within large language models. The paradigm established by PT underscores the potential for further breakthroughs at the intersection of planning and powerful sequence modeling architectures like transformers.

\bibliography{aaai25}

\appendix
\section*{Appendices}
\section{Experiment details}

\subsection{Environments}

We describe below each evaluation environment that we used to benchmark our model's performance:

\begin{enumerate}
    \item \textbf{Gym-Mujoco} \
    The Gym-Mujoco tasks involve a series of locomotion and control challenges with varying degrees of data optimality. We benchmark the medium-replay and expert categories, where medium-replay includes both medium and low-quality trajectories, while medium-expert includes medium and high-quality trajectories.
    
   There are three environments for each of these data types:
    \begin{enumerate}
        \item \textbf{HalfCheetah:} A simulated robot resembling a cheetah with the task of running forward as fast as possible. The robot has multiple joints and the control input includes forces applied at these joints.
        \item \textbf{Hopper:} A one-legged robot with the goal of hopping forward as far as possible without falling over.
        \item \textbf{Walker2d:} A bipedal robot that needs to maintain balance and walk forward as effectively as possible.
    \end{enumerate}

    As this is a purely reward-conditioned environment, we disable goal-conditioning by setting the goal to a 0-length vector.
    
    \item \textbf{AntMaze} \\
    The AntMaze tasks involve an 8-DOF quadruped ("Ant") robot, which must navigate various simulated mazes from a start to a goal. This environment is designed to test an RL agent's "trajectory stitching" abilities. It comes in four sizes: umaze, medium, large, ultra, where umaze is a U-shaped maze. Ultra is a larger and more challenging version of AntMaze proposed by \cite{TAP}
    
    Antmaze comes in two dataset qualities: play and diverse, where play is a handpicked selection of starts and goals, while diverse randomly picks starts and goals. We choose to benchmark only the more challenging diverse environments due to computational constraints. We expect that performance on Diverse is also reflective of performance on Play.

    Whilst Antmaze-Diverse is trained on random starts and goals trajectories. The evaluation is always with the start at the bottom left of the maze and the goal at the top right. We have found that this evaluation configuration results in performance superior than if the goal was randomly chosen in the maze due to a bias of the ant to reach the top right corner, but we maintain this evaluation procedure for consistency with other works.

    Goals are provided as a single x,y location. In training we use the first two indices of the observation to extract goals from trajectories.
    
    \item \textbf{FrankaKitchen} \\
    The FrankaKitchen environment involves a 9-DOF Franka robot performing various goal-conditioned kitchen tasks with a large observation space, with many objects that can be interacted with. The ultimate goal is to complete four tasks in any order. The tasks to complete depend on the specific environment:
    \begin{enumerate}
        \item \textbf{Partial}: (1) opening the microwave, (2) relocating the kettle, (3) toggling the light switch, and (4) initiating the sliding action of the cabinet door.
         \item \textbf{Mixed}: (1) opening the microwave, (2) relocating the kettle, (3) rotating the bottom burner knob, and (4) toggling the light switch.
    \end{enumerate} This is the most challenging environment due to its high observation space, complex series of tasks, and sparse rewards.

    We select the indices 11-29 from the State-Based observation space to use as the goal space. During evaluation, we use the observations with the goal indices selected and set the indices specific to the current tasks of the evaluation environment to their goal locations.
\end{enumerate}

\subsection{Evaluation Methodology}

To evaluate our models, we trained our model on the environment for a sufficient number of update steps, and then performed rollouts. The methodology used by \cite{rvs} is to train 5 random seeds and then perform 100 rollouts to get a mean normalized score. The average of these normalized scores and the standard deviation is reported. Ours is the same except we use 3 seeds instead of 5, due to computational limitations. This should not affect results significantly. It was not clear whether \cite{rvs} used the last checkpoint or the best checkpoint when scoring, but in interest of fairness and adhering to standard testing procedures in the field, we use last checkpoint.

\subsection{Selection of hyperparameters}
\begin{enumerate}
    \item For the FrankaKitchen and AntMaze experiments, we implement goal-conditioning with PT. For AntMaze, we restrict the Plans to only the first two features in the observation space, which contain the ant's body, as this provides a clear high-level observation space that can be used to make the Plans as effective as possible. However, for FrankaKitchen, there is no obvious subset of features to use as a high-level set, so we use all the features. For FrankaKitchen, we also included the actions.
    \item We removed the timestep embedding for all environments except for FrankaKitchen, as we hypothesised they may make policy generalisation more difficult to learn. For example, why should the route an ant takes be different whether it begins taking it after 0 steps or 500?
    \item For the Gym-MuJoCo tasks, which involve locomotion and have dense rewards, we apply reward-conditioning based on a target return. Unlike prior work, we don't constrain our high-level targets (the Plans) to only reward targets and instead use the full state, aswell as rewards to maximize the available information contained in the Plans.
    \item We found that target returns needed to be set to the maximum reward or slightly above it. We used 1.0 for AntMaze, 4.0 for Kitchen, and a normalised score of 110 for MuJoCo.
    \item For the AntMaze environment specifically, we found by visualizing the paths that a primary cause of failure was that the model would freeze up when the state went out-of-distribution. As a simple remedy, we added a small amount of noise to the action values, which was surprisingly effective.
\end{enumerate}

For the full set of hyperparameters used, please see Table \ref{Tab: hyperparams}.

\begin{table*}[h]
\centering
\caption{Hyperparameters and configuration details for PT across all experiments.}
\label{Tab: hyperparams}
\begin{tabular}{@{}lc@{}}
\toprule
Hyperparameter & Value \\
\midrule
Transformer Layers & 3 \\
Transformer Heads & 2 \\
Dropout Probability (attn) & 0.15 \\
Dropout Probability (resid) & 0.15 \\
Dropout Probability (embd) & 0.0 for Antmaze, 0.1 for others \\
Embedding Dimension & 192 Kitchen, 128 others \\
Non-Linearity & ReLU \\
Learning Rate & 0.0004 Mujoco, 0.0002 others \\
Gradient Update Steps & 200,000 Kitchen/Antmaze-Large/Antmaze-Ultra, 100,000 others \\
Batch Size & 256 Mujoco, 128 others \\
Sequence length & 20 Mujoco, 10 others \\
Timestep embedding? & True Kitchen, False others \\
Action noise scale & 35\% Antmaze-umaze/medium, 20\% Antmaze-large/ultra, 0 \% others \\
Max Trajectory Ratio & 1.0 Kitchen, 0.5 others \\
Fixed or Log plan sampling? & Log Antmaze, Fixed others \\
Include actions in plan? & True Kitchen, False others \\
Num. of Planning Tokens in Plan & 10 \\
Replanning-Interval ($\rho$) & 10 \\
\bottomrule
\end{tabular}
\end{table*}

\end{document}